\pdfoutput=1

\documentclass[11pt]{article}

\usepackage{acl}

\usepackage{times}
\usepackage{latexsym}

\usepackage[T1]{fontenc}

\usepackage[utf8]{inputenc}

\usepackage{microtype}

\usepackage{inconsolata}
\usepackage{seqsplit}
\usepackage{autobreak}
\usepackage{mathtools}
\usepackage{url}
\usepackage{adjustbox}
\usepackage{rotating}
\usepackage{booktabs}
\usepackage{multirow}
\usepackage{comment}
\DeclareMathOperator*{\argmax}{arg\,max}

\title{Can we obtain significant success in RST discourse parsing\\ by using Large Language Models?}

\newcommand{\mail}[2]{\href{mailto:#1}{\color{black}{#2}}}
\author{
    \textbf{Aru Maekawa}\textsuperscript{\rm 1},
    \textbf{Tsutomu Hirao}\textsuperscript{\rm 2},
    \textbf{Hidetaka Kamigaito}\textsuperscript{\rm 1},
    \textbf{Manabu Okumura}\textsuperscript{\rm 1}\\
    \textsuperscript{\rm 1}Institute of Innovative Research, Tokyo Institute of Technology,\\
    \textsuperscript{\rm 2}NTT Communication Science Laboratories, NTT Corporation\\
    \texttt{\{\mail{maekawa@lr.pi.titech.ac.jp}{maekawa@lr.},
              \mail{kamigaito@lr.pi.titech.ac.jp}{kamigaito@lr.},
              \mail{oku@pi.titech.ac.jp}{oku@}%
            \}pi.titech.ac.jp}\\
    \texttt{\mail{tsutomu.hirao@ntt.com}{tsutomu.hirao@ntt.com}} 
}

\begin{document}
\maketitle
\begin{abstract}
Recently, decoder-only pre-trained large language models (LLMs), with several tens of billion parameters, have significantly impacted a wide range of natural language processing (NLP) tasks.
While encoder-only or encoder-decoder pre-trained language models have already proved to be effective in discourse parsing, the extent to which LLMs can perform this task remains an open research question.
Therefore, this paper explores how beneficial such LLMs are for Rhetorical Structure Theory (RST) discourse parsing.
Here, the parsing process for both fundamental top-down and bottom-up strategies is converted into 
prompts, which LLMs can work with. 
We employ Llama~2 and fine-tune it with QLoRA, which has fewer parameters that can be tuned.
Experimental results on \textcolor{black}{three benchmark datasets, RST-DT, Instr-DT, and the GUM corpus,} demonstrate that Llama~2 with 70 billion parameters in the bottom-up strategy obtained state-of-the-art (SOTA) results with significant differences. 
\textcolor{black}{
Furthermore, our parsers demonstrated generalizability when evaluated on RST-DT, showing that, in spite of being trained with the GUM corpus, 
it obtained similar performances to those of existing parsers trained with RST-DT. 
}
\end{abstract}

\section{Introduction}

\begin{figure}[t]
    \centering
    \includegraphics[width=\linewidth]{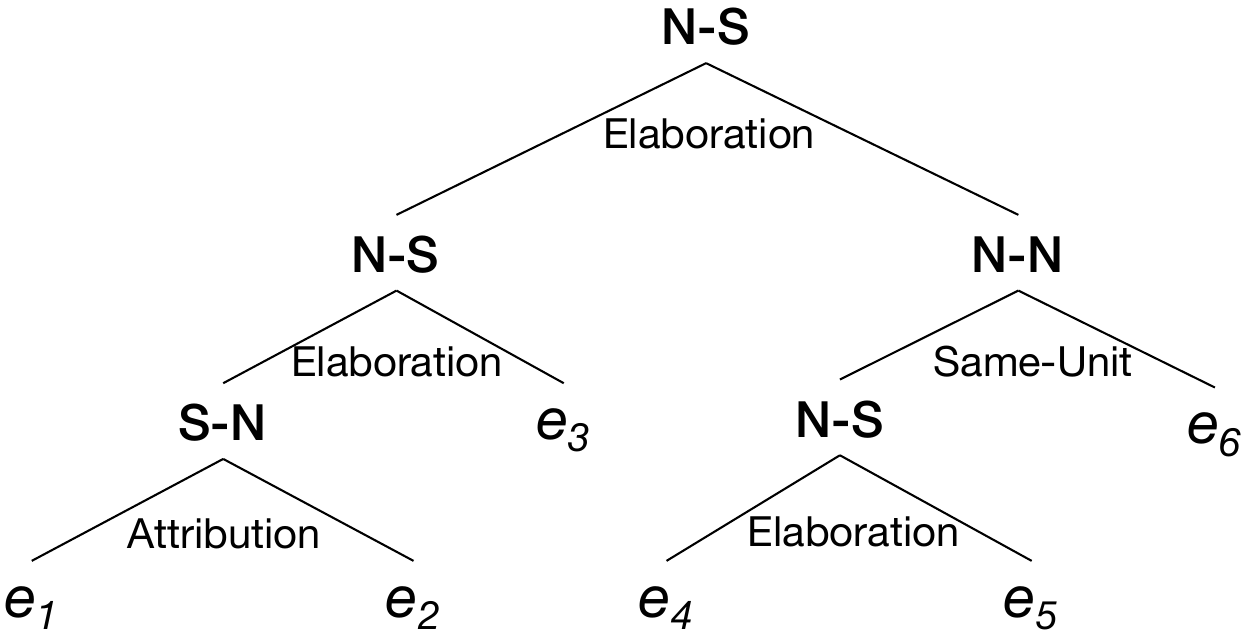}
    \caption{
     Example of RST tree from \texttt{WSJ\_1100} in RST-DT \cite{carlson-etal-2002-rstdt}, consisting of six EDUs ($e$): 
        $e_1$: [Westinghouse Electric Corp. said],
        $e_2$: [it will buy Shaw-Walker Co.],
        $e_3$: [Terms weren't disclosed.],
        $e_4$: [Shaw-Walker,],
        $e_5$: [based in Muskegon, Mich.,],
        $e_6$: [makes metal files and desks, and seating and office systems furniture.]. N and S represent the Nucleus and Satellite, respectively.
        }
    \label{fig:rst}
\end{figure}

Rhetorical Structure Theory (RST) \cite{mann:87:a} is one of the influential discourse theories used to explain the coherence of texts. It plays an important role in various natural language processing (NLP) tasks at the document level, including sentiment analysis \cite{bhatia-etal-2015-better}, automatic summarization \cite{marcu-1998-improving,xu-etal-2020-discourse,kwon2021}, question answering \cite{gao-etal-2020-discern}, machine translation \cite{chen-etal-2020-modeling,ijcai2022p608}, and MT evaluation \cite{joty-etal-2017-discourse}. According to RST, a text is represented as a binarized constituent tree (RST-tree), whose terminal nodes correspond to elementary discourse units (EDUs), clause-like units, and non-terminal nodes indicate the nuclearity status, i.e., either Nucleus or Satellite, of text spans consisting of single or contiguous EDUs. The edges represent the rhetorical relation between two adjacent text spans dominated by non-terminal nodes. 
Figure \ref{fig:rst} shows an example of the RST tree obtained from RST Discourse Treebank (RST-DT) \cite{carlson-etal-2002-rstdt}. In the figure, the nucleurity status of the text span consisting of $e_1$ and $e_2$ is the nucleus, and it is modified by the satellite, $e_3$. A mono-nuclear relation, Attribution, is given to the two spans. 

\textcolor{black}{
Since the late 2010s, neural RST discourse parsing methods that use encoder-only pre-trained language models (PLMs) to encode text spans into vectors have been proposed due to advances in neural models.}
While earlier models, e.g., \citet{yu-etal-2018-transition,lin-etal-2019-unified,kobayashi-etal-2020-topdown}, obtained vector representations for text spans from a static PLM, such as GloVe \cite{pennington-etal-2014-glove}, recent models, e.g., \citet{guz-carenini-2020-coreference, shi2020endtoend,nguyen-etal-2021-rst, zhang-etal-2021-adversarial}, obtained them from a transformer-based PLM, such as XLNet \cite{NEURIPS2019_dc6a7e65}.
\textcolor{black}{
To form RST trees, they obtained vectors via PLMs and exploited them to determine the parsing actions in a top-down or bottom-up strategy.}
More recently, there is also a parser \cite{10224326} that utilizes an encoder-decoder PLM to transform input text into a linearized RST tree.

\textcolor{black}{
There was a shift in focus from encoder-only to massive-scale decoder-only PLMs. }
Some large language models (LLMs), such as GPT-3 \cite{NEURIPS2020_1457c0d6} and Llama~2 \cite{touvron2023llama}, have several tens of billions of parameters and are pre-trained with only a decoder.
These have significantly impacted NLP, similar to encoder-only and encoder-decoder PLMs.
LLMs have demonstrated remarkable success in various NLP tasks due to their large numbers of parameters and ease of availability. 
Their impact extends beyond generation tasks and includes classification tasks \cite{NEURIPS2020_1457c0d6,weifinetuned,wu2023exploring}. 
Therefore, they could also be advantageous in RST discourse parsing.
Furthermore, we are strongly motivated to adopt LLMs because previous discourse parsing methods have been greatly improved by using 
encoder-only or encoder-decoder pre-trained language models.

In this paper, we explore the potential of using LLMs for RST discourse parsing. 
As a first step in exploiting LLMs, our approach is to translate the parsing steps of both fundamental top-down and bottom-up strategies into prompts.
Then, we fine-tune Llama~2 using QLoRA \cite{dettmers2023qlora}, an extension of LoRA \cite{hu2022lora}, which is an adapter that injects trainable low-rank matrices into each layer of the Transformer, while it freezes the weights of the pre-trained model for efficient computing.
The experimental results from RST-DT, Instructional Discourse Treebank (Instr-DT) \cite{subba-di-eugenio-2009-effective}, and the GUM corpus \cite{UD_GUM:2017} demonstrate that our parser with the bottom-up parsing strategy surpassed the current state-of-the-art (SOTA) results. 
\textcolor{black}{
It outperformed the current SOTA models 
by around 2-3 points on RST-DT, by 0.4-3.7 points on Instr-DT, and by 1.5-6 points on the GUM corpus.}
\textcolor{black}{
Furthermore, out-of-domain evaluations using RST-DT and the GUM corpus demonstrate the potential generalizability of our parsers. 
Our parsers, trained with the GUM corpus, achieved smaller degradation when evaluated on RST-DT. The performances are close to those of the existing 
parsers trained with RST-DT itself.
These findings provide valuable insights into the future direction of RST discourse parsing.
}
We will release our code at \url{https://github.com/nttcslab-nlp/RSTParser_EACL24}.

\section{Related Work}

\subsection{RST Discourse Parsing with Encoder-only PLMs}

Most neural RST discourse parsers have two fundamental components: a feature extraction layer to obtain vector representations for text spans and a classification layer to form RST trees. 
The feature extraction layer receives tokens in the text spans as an input and obtains their vector representations through a PLM.
The classification layer, located on top of the feature extractor, makes decisions that guide the form of RST trees by the parsers. 

\citet{yu-etal-2018-transition} proposed a bottom-up parsing model with a feature extractor based on GloVe 
and syntactic features. The parser merges text spans using shift-reduce operations to build RST trees based on Feed-Forward Networks (FFNs). To extend the parser, they incorporated a BERT-based tailored PLM with objectives that include the prediction of the next EDU and a discourse marker \cite{yu-etal-2022-rst}.
By enhancing the PLM, the performance was greatly improved: They achieved a fully-labeled span F1 score of 53.8.
\citet{guz-carenini-2020-coreference} extended Wang et al.'s classical shift-reduce parser \cite{wang-etal-2017-two}, replacing SVMs with FFNs and a feature extractor with SpanBERT \cite{joshi-etal-2020-spanbert}. 
The gain of F1 scores against Wang et al.’s parser was around 3 points due to having more sophisticated contextual word embeddings.

\citet{kobayashi-etal-2020-topdown} proposed a top-down parsing model based on a minimal span-based approach, that recursively splits a span into smaller ones by exploiting a classification layer with FFNs.
Their feature extractor was a combination of Glove and ELMo \cite{peters-etal-2018-deep}.
Another top-down parsing model was proposed 
 using a decoder instead of FFNs for a classification layer.
\citet{lin-etal-2019-unified} proposed top-down depth-first parsing at the sentence-level based on a pointer-generator network. The parser employs GloVe in the feature extractor, and then the decoder recursively generates a split for an input span.
\citet{shi2020endtoend} introduced layer-wise beam search and used XLNet \cite{NEURIPS2019_dc6a7e65} in the feature extractor to extend the top-down model to the document level, achieving SOTA 
results at that time.
\citet{nguyen-etal-2021-rst} and \citet{zhang-etal-2021-adversarial} also reported that XLNet is beneficial for enhancing performance in a similar top-down approach. 

Recently, \citet{kobayashi-etal-2022-simple} explored simple and strong baselines based on Guz and Carenini's bottom-up parser \citeyearpar{guz-carenini-2020-coreference} and Kobayashi et al.'s top-down parser \citeyearpar{kobayashi-etal-2020-topdown} with varying encoder-only PLMs.
The results suggest that the success of the parsing method heavily relies on the PLMs rather than on the parsing strategies themselves. 
The current best score, a fully-labeled span score of 55.4, was obtained by the bottom-up parser combined with DeBERTa \cite{he2021deberta}, a SOTA encoder-only PLM.

As another approach, \citet{braud-etal-2016-multi} proposed RST discourse parsing as a text-to-text generation task.\footnote{While this approach may seem appropriate to the next section, since the studies exclusively used only encoder-only PLMs, we categorize them here.}
They used an LSTM-based encoder-decoder to receive a text as an input and output an S-expression that expresses a path from a root node to leaf nodes of an RST tree. They adopted an earlier PLM, PolyGlot \cite{al-rfou-etal-2013-polyglot}.
\citet{zhang-etal-2021-language} proposed sentence-level parsing by re-ranking linearized parse trees obtained from an external parser, based on MPNet \cite{NEURIPS2020_c3a690be}. 

\subsection{RST Discourse Parsing with Encoder-decoder PLMs}


As an extension of Braud et al.'s approach \citeyearpar{braud-etal-2016-multi}, 
\citet{10224326} proposed a more straightforward model as a text-to-text generation task, that exploits a SOTA encoder-decoder PLM, T5 \cite{JMLR:v21:20-074}.
It directly learns the transformation from an input text into the linearized S-expression of the entire RST tree by benefiting from a powerful encoder-decoder PLM.
The parser performed better than Nguyen et al. \citeyearpar{nguyen-etal-2021-rst}.



\section{Proposed Approach}


Three approaches can be possible when using decoder-only LLMs for RST
discourse parsing. The first is to create linearized S-expressions for RST trees using LLMs, which is similar to how encoder-decoder PLMs have been used.
\textcolor{black}{
The second is to replace encoder-only PLMs with an encoder of LLMs in conventional top-down or bottom-up parsing.}
The third method involves using LLMs to imitate the parsing process, which involves combining feature extraction and classification layers, similar to how encoder-only PLMs have been used. 
\textcolor{black}{
The first approach can be challenging, especially when dealing with lengthy documents. This is because the number of output tokens increases disproportionately with the number of input tokens. Furthermore, additional techniques, such as constrained beam search, are required to obtain valid linearization forming a tree.
The second approach is not promising because it does not perform as well as encoder-only PLMs \cite{devlin-etal-2019-bert}.
Thus, we adopt the third approach in this work.
}


Before describing our own approach, we first illustrate fundamental top-down and bottom-up parsing methodologies using encoder-only PLMs. 

\subsection{Bottom-up Parsing}
\label{sec:bu}

\begin{figure}[t]
    \centering
    \includegraphics[width=\linewidth]{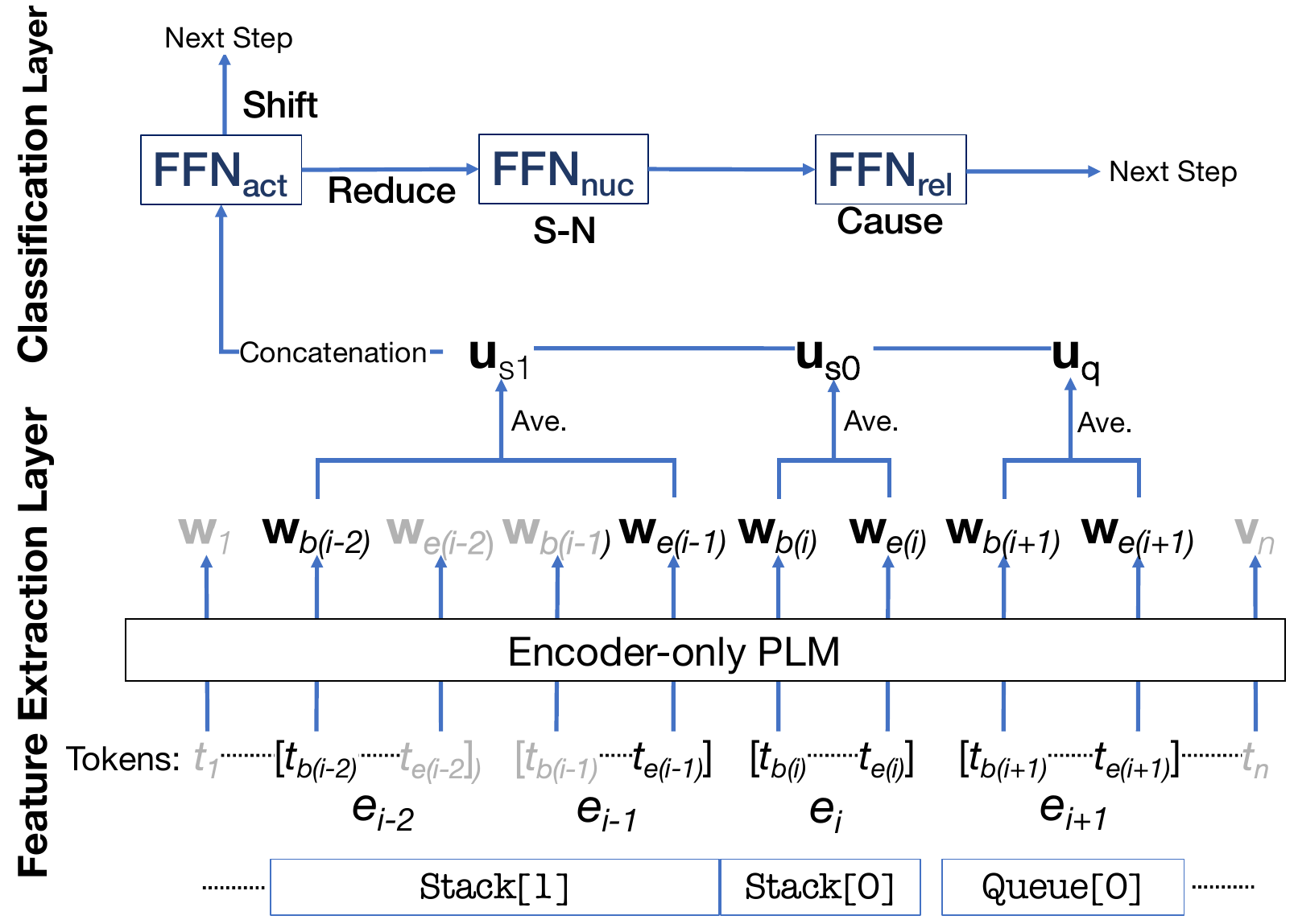}
    \caption{Bottom-up parsing by shift-reduce operations}
    \label{fig:BU}
\end{figure}

Figure \ref{fig:BU} gives an overview of bottom-up parsing based on shift-reduce algorithms.
The text's tokens are first converted into word embeddings using an encoder-only PLM. Then, a vector representation for a text span is obtained by averaging the word embeddings for 
the leftmost token in the first EDU and the rightmost token in the final EDU.

In the figure, FFNs in the classification layer handles shit-reduce operations based on a stack and a queue; 
a stack stores subtrees, i.e., text spans that have already been parsed, and a queue contains incoming EDUs.
The parser builds an RST tree 
by merging two adjacent text spans while selecting one of the following actions:
\begin{description}
\item{Shift}: Pop the first EDU off the queue and push it onto the stack.
\item{Reduce}: Pop two text spans from the stack and merge them into a new span, then push it onto the stack.
\end{description}
Note that the nuclearity status and rhetorical relation labels are independently predicted by different classifiers when the Reduce operation is selected. 
$\textrm{FFN}_{\rm act}$, $\textrm{FFN}_{\rm nuc}$, and $\textrm{FFN}_{\rm rel}$ in the classification layer are feed-forward networks for predicting the action, nuclearity, and relation labels, respectively. 
$\textrm{FFN}_{\rm act}$ solves a binary classification problem (Shift or Reduce), $\textrm{FFN}_{\rm nuc}$
solves a three-class classification problem (either nucleus-nucleus, nucleus-satellite, or satellite-nucleus), and $\textrm{FFN}_{\rm rel}$ solves
a multi-class classification problem (the number of classes derives from the number of rhetorical relations used in the dataset): 
    $s_{\rm *} {=} \textrm{FFN}_{\rm *}(
        \textrm{Concat}(\mathbf{u}_{{\rm s}_0}, \mathbf{u}_{{\rm s}_1}, \mathbf{u}_{{\rm q}_0})),$
where the function ``Concat'' concatenates the vectors received as the arguments.
$\mathbf{u}_{{\rm s}_0}$ is the vector representation of a text span stored in
the first position of the stack,
$\mathbf{u}_{{\rm s}_1}$ is that in the second position,
and $\mathbf{u}_{{\rm q}_0}$ is that in the first position of the queue.

\subsection{Top-down Parsing}
\label{sec:td}

\begin{figure}[t]
    \centering
    \includegraphics[width=\linewidth]{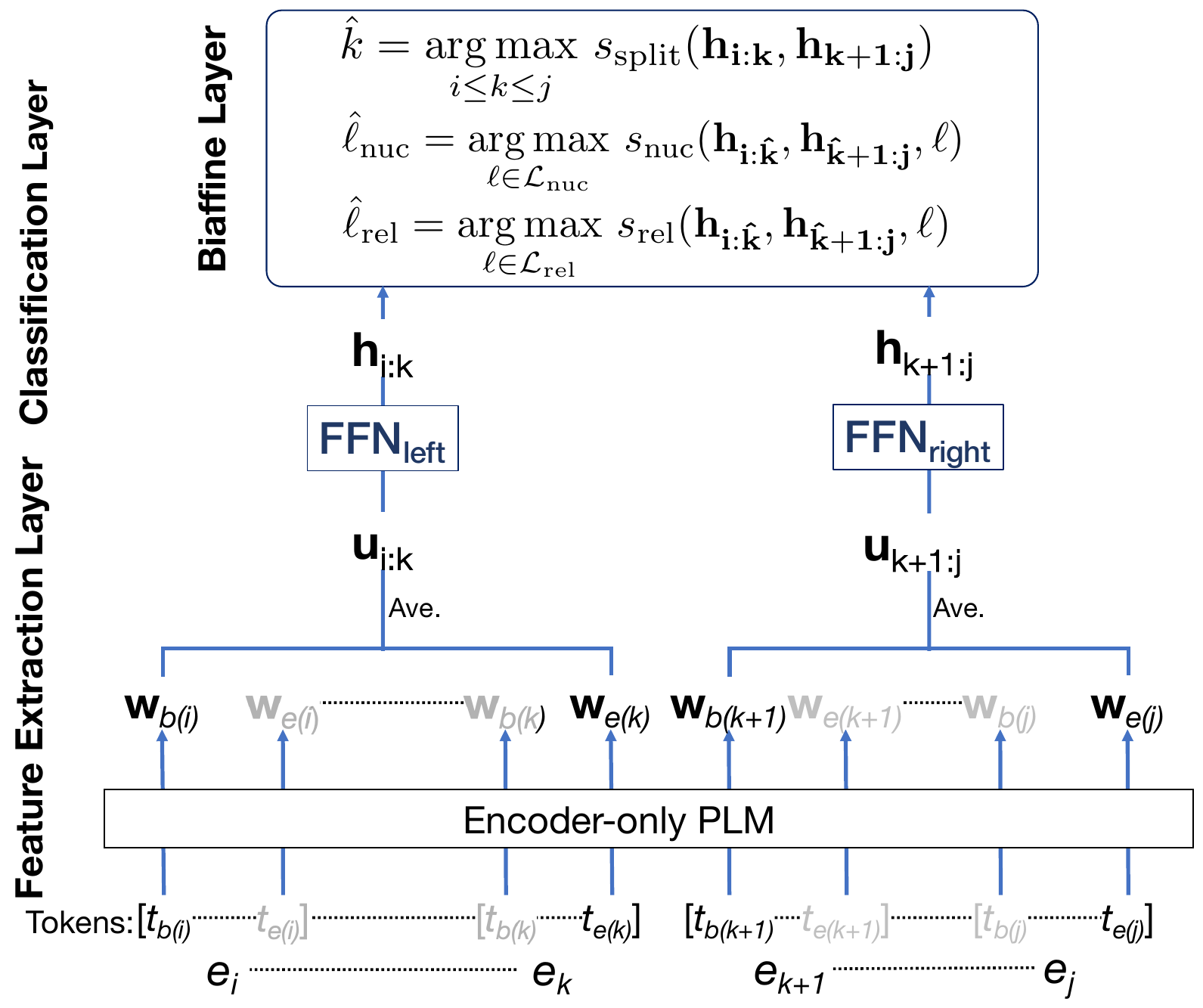}
    \caption{Top-down parsing by span split
        }
    \label{fig:TD}
\end{figure}

An overview of top-down parsing is presented in Figure \ref{fig:TD}. 
We obtain two vector representations of text spans, $\mathbf{u}_{i:k}$ and $\mathbf{u}_{k{+}1:j}$, for each possible split point $k$ of the span between the $i$-th and $j$-th EDUs, using the same approach as in the bottom-up parsing.
Then, the classification layer consisting of FFNs and the biaffine layer identify the best-split point based on a scoring function, $s_{\rm split}(\mathbf{h}_{i:k}, \mathbf{h}_{k{+}1:j})$, which is defined as
\begin{align}
    \begin{autobreak}
    s_{\text{split}}(\mathbf{h}_{i:k}, \mathbf{h}_{k{+}1:j}) = 
        \mathbf{h}_{i:k}\mathbf{W} \mathbf{h}_{k+1:j}
        + \mathbf{v}_{\rm left} \mathbf{h}_{i:k}
        + \mathbf{v}_{\rm right} \mathbf{h}_{k+1:j}, 
    \end{autobreak}
     \label{split_score}
\end{align}
where $\mathbf{W}$ is a weight matrix and $\mathbf{v}_{\rm left}$ and $\mathbf{v}_{\rm right}$ are weight vectors corresponding to the left and right spans, respectively.
Here, $\mathbf{h}_{i:k}$ and  $\mathbf{h}_{k+1:j}$ are obtained via FFNs
 as follows:
\begin{align}
    \mathbf{h}_{i:k} &= \textrm{FFN}_{\text{left}}(\mathbf{u}_{i:k}), \\
    \mathbf{h}_{k{+}1:j} &= \textrm{FFN}_{\text{right}}(\mathbf{u}_{k{+}1:j}).  
    \label{eq:td_split}
\end{align}
Then, the span is split at the position $k$ that maximizes Eq.~(\ref{split_score}). 

For a pair of text spans divided at point $\hat{k}$, we assign either nucleus-nucleus, nucleus-satellite, or satellite-nucleus and a rhetorical relation from a pre-defined set
using the following scoring function: 
\begin{align}
    \begin{autobreak}
    s_{\text{label}}(\mathbf{h}_{i:\hat{k}}, \mathbf{h}_{\hat{k}{+}1:j},\ell) = 
        \mathbf{h}_{i:\hat{k}}\mathbf{W}^{\ell} \mathbf{h}_{\hat{k}+1:j}
        {+} \mathbf{v}_{\rm left}^{\ell} \mathbf{h}_{i:\hat{k}}
        {+} \mathbf{v}_{\rm right}^{\ell} \mathbf{h}_{\hat{k}+1:j},  
    \end{autobreak}
     \label{label_score}
\end{align}
where $\mathbf{W}^{\ell}$ is a weight matrix for a specific nuclearity or relation label $\ell$ and 
 $\mathbf{v}_{\rm left}^{\ell}$ and $\mathbf{v}_{\rm right}^{\ell}$
 are weight vectors corresponding to the left and right spans for the label $\ell$, respectively.



\subsection{Prompts for Bottom-up Parsing}

\begin{figure*}[t]
    \centering
    \includegraphics[width=\linewidth]{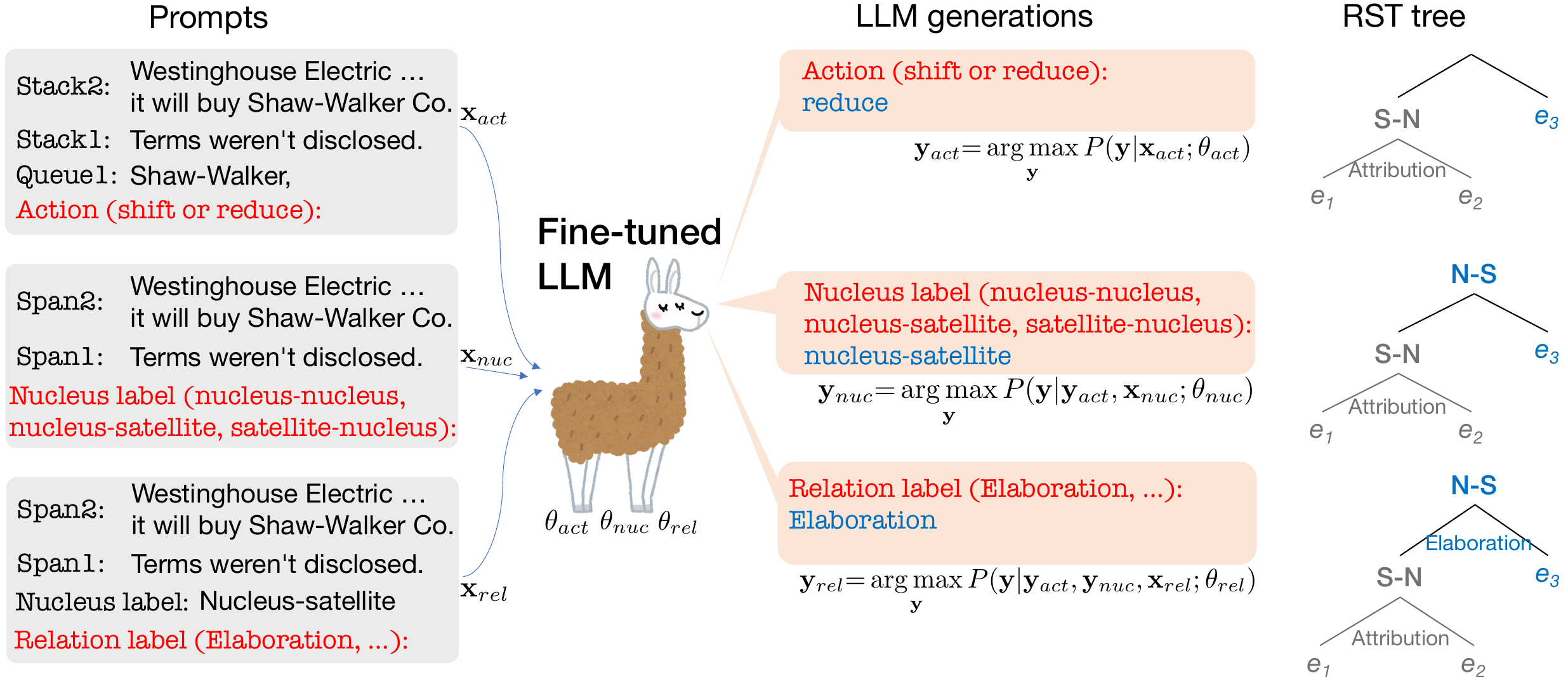}
    \caption{Example of the bottom-up parsing process using an LLM with prompts. In the example, \texttt{Stack2} stores a text span, an already constructed subtree, consisting of two EDUs:   
   $e_1$: [Westinghouse Electric Corp. said],
   $e_2$: [it will buy Shaw-Walker Co.]. \texttt{Stack1} stores a text span of single EDU,
   $e_3$: [Terms weren't disclosed.]. \texttt{Queue1} stores an EDU,
   $e_4$: [Shaw-Walker,]. After this step, the parsing process goes on to the next steps while updating Stack* and Queue.}
    \label{fig:BULLM}
\end{figure*}

To parse a document in a bottom-up manner with LLMs, we translate the shift-reduce operations, described in \S \ref{sec:bu}, into prompts. 
In this case, we emulate the parsing process with a stack and a queue while using the following prompt template $\mathbf{x}_{act}$ to predict an action $\mathbf{y}_{act}$:
\begin{quote}
    \textbf{\texttt{Stack2}}: \textit{Text span(s) in the second position of the stack.}\\
    \textbf{\texttt{Stack1}}: \textit{Text span(s) in the first position of the stack.}\\
    \textbf{\texttt{Queue1}}: \textit{An EDU in the first position of the queue.}\\
    \textbf{\texttt{Action (shift or reduce)}}: \textit{Either Shift or Reduce}.
\end{quote}
An LLM determines whether to Shift or Reduce based on the text spans in Stack1, Stack2, and Queue1 at each step with the above prompts.
Then, we assign nuclearity and rhetorical relation labels between two text spans in Stack1 and Stack2 when the Reduce action is selected. 
We use the following prompt template $\mathbf{x}_{nuc}$ to predict a nuclearity label $\mathbf{y}_{nuc}$:\\
\begin{quote}
    \textbf{\texttt{Span2}:} \textit{Text span(s) in the second position of the stack.}\\
    \textbf{\texttt{Span1}:} \textit{Text span(s) in the first position of the stack.}\\
    \textbf{\texttt{Nucleus label (\seqsplit{nucleus-nucleus}, \seqsplit{nucleus-satellite}, \seqsplit{satellite-nucleus})}:} \textit{Either one of them.}
\end{quote}
Here, Span1 and Span2 are text spans in Stack1 and Stack2, respectively. 
To predict a rhetorical relation label $\mathbf{y}_{rel}$, we use the prompt template $\mathbf{x}_{rel}$ by replacing the third prompt of $\mathbf{x}_{nuc}$ with the following two new prompts:
\begin{quote}
    \textbf{\texttt{Nucleus label}}: \textit{predicted nuclearity label.}\\
    \textbf{\texttt{Relation label (Rel$_1$, Rel$_2$, $\ldots$, Rel$_n$)}}: \textit{Either one of them.}
\end{quote}
Here, Rel$_i$ indicates the $i$-th rhetorical relation, and the set of rhetorical relations is different for each dataset.
We construct an RST tree based on LLM's decisions using the above prompts for each parsing step.
Figure \ref{fig:BULLM} shows an example of our bottom-up parsing with prompts.

\textcolor{black}{LLMs infer outputs by choosing the sequence with the maximum probability as follows:}
\begin{align}
\!\!\!&\mathbf{y}_{act}=\textstyle\argmax_{\mathbf{y}}P(\mathbf{y}|\mathbf{x}_{act};\mathbf{\theta}_{act}),\\
\!\!\!&\mathbf{y}_{nuc}=\textstyle\argmax_{\mathbf{y}}P(\mathbf{y}|\mathbf{y}_{act}, \mathbf{x}_{nuc};\mathbf{\theta}_{nuc}),\\
\!\!\!&\mathbf{y}_{rel}=\textstyle\argmax_{\mathbf{y}}P(\mathbf{y}|\mathbf{y}_{act},\mathbf{y}_{nuc}, \mathbf{x}_{rel};\mathbf{\theta}_{rel}),
\end{align}
\textcolor{black}{where $\mathbf{\theta}_{act}$, $\mathbf{\theta}_{nuc}$, and $\mathbf{\theta}_{rel}$ are weight parameters of LLMs for predicting action, nuclearity, and rhetorical relation labels, respectively.}

\subsection{Prompts for Top-down Parsing}

\begin{figure*}[t]
    \centering
    \includegraphics[width=\linewidth]{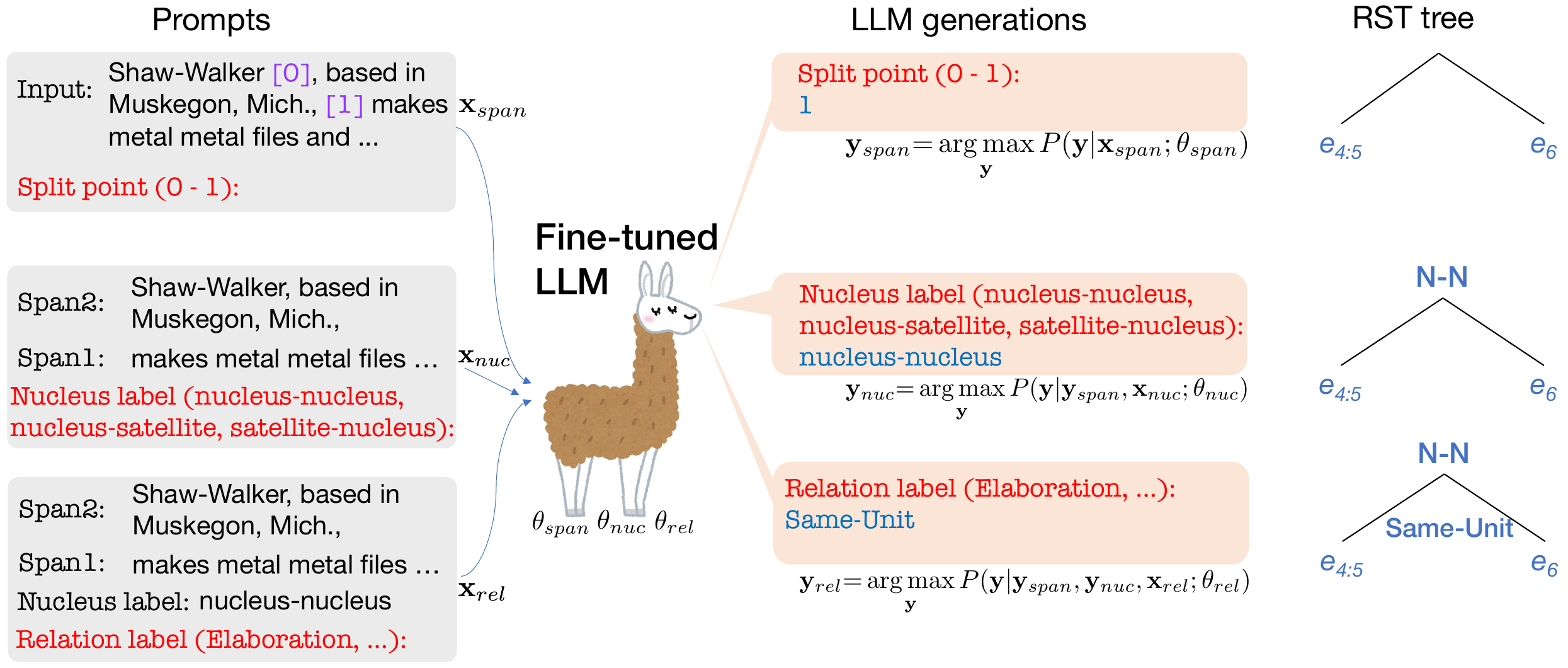}
    \caption{Example of the top-down parsing process using an LLM with prompts. In the example, a text span consisting of three EDUs,   
   $e_4$: [Shaw-Walker,],
   $e_5$: [based in Muskegon, Mich.,], and 
   $e_6$: [makes metal files and desks, and seating and office systems furniture.],
   are divided and labeled by an LLM's decision.
   The process is recursively applied until divided spans are identical to single EDUs.}
    \label{fig:TDLLM}
\end{figure*}

To perform top-down parsing with LLMs, 
we translate the span split procedure, described in \S\ref{sec:td}, into the following prompt template $\mathbf{x}_{span}$ for predicting a split span $\mathbf{y}_{span}$:
\begin{quote}
\textbf{\texttt{Input}:} \textit{A text span, a sequence of EDUs, to be split.}\\
\textbf{\texttt{Split point}} ($0$ –- $j{-}i{-}1$): \textit{An index of EDU ($0 \le k \le j{-}i{-}1$).}
\end{quote}
Note that we adjust the index of EDUs in the given text span, consisting of the $i$-th EDU to the $j$-th EDU, so that the index of the first EDU in the prompt is always 0.
We give the above prompts recursively to an LLM in order to identify a split point for a span obtained during parsing.

We use the same prompts as in bottom-up parsing to assign nuclearity and rhetorical relation labels.
Figure \ref{fig:TDLLM} shows an example.

\textcolor{black}{Similar to the bottom-up parsing, inference is performed as follows:}
\begin{align}
\!\!\!&\mathbf{y}_{span}=\textstyle\argmax_{\mathbf{y}}P(\mathbf{y}|\mathbf{x}_{span};\mathbf{\theta}_{span}),\\
\!\!\!&\mathbf{y}_{nuc}=\textstyle\argmax_{\mathbf{y}}P(\mathbf{y}|\mathbf{y}_{span}, \mathbf{x}_{nuc};\mathbf{\theta}_{nuc}),\\
\!\!\!&\mathbf{y}_{rel}\!=\!\textstyle\argmax_{\mathbf{y}}P(\mathbf{y}|\mathbf{y}_{span},\!\mathbf{y}_{nuc},\! \mathbf{x}_{rel};\!\mathbf{\theta}_{rel}),
\end{align}
\textcolor{black}{where $\mathbf{\theta}_{span}$ denotes weight parameters for the LLM to split spans.}

\subsection{Handling Erroneous Generation}

Either of our parsing approaches might generate labels not in a pre-defined set for the classification.
Such labels prevent the construction of valid RST trees.
Accordingly, we introduce default rules to correct such invalid generation. When invalid generation occurs, one of the following rules is applied and the default label is used in place of the generated one: 
Shift is the default action for bottom-up parsing, 0 is the default split point for top-down parsing. Nucleus-satellite and Elaboration are the default nuclearity and rhetorical relation labels, respectively, for both parsing strategies.
These are selected because of 
the majority labels in our datasets. 

\section{Experimental Settings}

\subsection{Large Language Models}

We utilized Llama~2 \cite{touvron2023llama},\footnote{
\url{https://huggingface.co/meta-llama/Llama-2-{7,13,70}b-hf}
} one of the largest publicly available open-source decoder-only PLMs, for our LLM-based RST discourse parsing models. 
Llama~2 can handle up to 70 billion parameters, which have been trained using a trillion tokens of text data. Although the technical report did not provide information on the specifics of the training dataset, it was made clear that the dataset comprises publicly available sources. Hence, we are confident that the datasets used for training and testing our parsers are not included in Llama~2.

Since we have found that zero-shot and few-shot approaches do not produce satisfactory results,\footnote{
\textcolor{black}{
A zero-shot approach resulted in only a 9.27 Span F-score, indicating no further consideration. This finding is expected because LLM pre-training did not cover parsing actions like shift or reduce. Since LLMs lack this inherent knowledge due to their pre-training, fine-tuning them emerges as the most viable approach.}
} 
we instead opted to fine-tune Llama~2 with the prompts and the correct outputs.
However, the large GPU memory requirements and computational costs made it impractical to fully fine-tune Llama~2. As a result, we turned to QLoRA \cite{dettmers2023qlora}\footnote{\url{https://github.com/artidoro/qlora}} to \textcolor{black}{update $\mathbf{\theta}_{act}$, $\mathbf{\theta}_{span}$, $\mathbf{\theta}_{nuc}$, and $\mathbf{\theta}_{rel}$}. QLoRA is a quantized version of LoRA \cite{hu2022lora}, which introduces trainable low-rank matrices into each layer of the Transformer architecture 
without altering the weights for the parameters in LLMs. 
\renewcommand{\arraystretch}{0.8}
\setlength{\tabcolsep}{4pt}
\begin{table*}[t]
    \centering
    \begin{tabular}{llcccccccccccc}
    \toprule
    & & \multicolumn{4}{c}{\textbf{RST-DT}} & \multicolumn{4}{c}{\textbf{Instr-DT}} & \multicolumn{4}{c}{\textbf{GUM}}\\       
    \cmidrule(lr){3-6}\cmidrule(lr){7-10}\cmidrule(lr){11-14}
    & & \textbf{Span} & \textbf{Nuc.} & \textbf{Rel.} & \textbf{Full} & 
    \textbf{Span} & \textbf{Nuc.} & \textbf{Rel.} & \textbf{Full} &
    \textbf{Span} & \textbf{Nuc.} & \textbf{Rel.} & \textbf{Full}
    \\
    \midrule
        \multirow{6}{*}{\rotatebox[origin=c]{90}{{Top-down}}}
    & Liu et al.
    & 76.5 & 65.2 & 54.2 & $-$ &$-$ &$-$ &$-$ &$-$ & 68.6 & 54.9 & $-$ & $-$ \\
    & Yu et al. & 72.9 & 62.7 & 52.5 & 50.5 & $-$ &$-$  &$-$  & $-$ &$-$ &$-$ &$-$ & $-$\\
    & Kobayashi et al. 
    & 78.5 &  67.9 & 56.6 & 54.4 & 77.3 & 57.9 & 50.0 & 43.4
    & 74.4 & 62.2 & 50.9& 48.7\\
    & Llama~2 (7B) & 76.3 & 65.4 & 55.2 & 53.4 & 75.7 & 56.2 & 49.8 & 43.6
    & 72.8& 60.9& 52.1& 50.9
    \\
    & Llama~2 (13B) & 78.6 & 67.9 & 57.7 & 55.6 & 75.7 & 57.3 & 50.2 & 43.6
    & 74.9& 62.5& 53.8&52.5
    \\
    & Llama~2 (70B) & 78.8 & 68.7 & 57.7 & 56.0 & 76.2 & 57.1 & 53.1 & 45.2
    & 75.8 &64.0& 55.8&54.8
    \\
    \midrule
    \multirow{6}{*}{\rotatebox[origin=c]{90}{{Bottom-up}}}
    & Guz et al.
    & 76.5 & 65.9 & 54.8 & $-$ &$-$ &$-$ &$-$ &$-$ & 69.9 & 57.0 & $-$ & $-$ \\
    & Yu et al. 
    & 76.4 & 66.1 & 54.5 & 53.5 & $-$ &$-$  &$-$  & $-$ &$-$ &$-$ &$-$ & $-$\\
    & Kobayashi et al. 
    & 77.8 & 68.0 & 57.3 & 55.4 & 77.8 & 60.0 & 51.4 & 44.4
    & 73.4& 60.9& 50.3& 48.5
    \\ 
    & Llama~2 (7B) & 78.2 & 67.5 & 57.6 & 55.8 & 76.7 & 58.2 & 48.5 & 43.5
    &74.4&63.0& 53.4& 52.1
    \\
    & Llama~2 (13B) & 78.3 & 68.1 & 57.8 & 56.0 & 77.4 & 60.4 & 52.1 & 46.1
    & 74.8& 63.4& 54.0&52.8    
    \\
    & Llama~2 (70B) & \textbf{79.8} & \textbf{70.4} & \textbf{60.0} & \textbf{58.1} & \textbf{79.1} & \textbf{60.4} & \textbf{55.1} & 
    \textbf{47.3} 
    & \textbf{76.4}& \textbf{64.7}& \textbf{56.4}&\textbf{55.2}    
    \\
    \bottomrule
    \end{tabular}
    \caption{Results on RST-DT, Instr-DT, and the GUM Corpus with Standard-Parseval. \textcolor{black}{Liu et al.'s top-down parser \cite{liu-etal-2021-dmrst} employed XLM-RoBERTa-base with 125M parameters, Guz et al's bottom-up parser \cite{guz-carenini-2020-coreference} employed SpanBERT-base with 110M parameters, Yu et al.'s parsers \cite{yu-etal-2022-rst} employed XLNet-base with 110M parameters, and Kobayahi et al.'s parsers employed DeBERTa-base with 140M parameters.
    We omit the reported Relation scores for Liu et al.'s and Guz et al.'s parsers on the GUM corpus because they are the results based on GUM's own relation label set.}
    }
    \label{tab:results}
\end{table*}
\setlength{\tabcolsep}{6pt}

\subsection{Datasets}

LLMs are pre-trained on large open-domain datasets, allowing our parsers to easily adapt to specific domains through fine-tuning.
To demonstrate this capability and make a fair comparison with Kobayashi et al.'s bottom-up and top-down parsers, we used two benchmark datasets from different domains,
RST-DT, Instr-DT, and GUM Corpus.

RST-DT contains 385 documents selected from the Wall Street Journal.
It is officially divided into 347 documents as the training set and 38 as the test dataset. 
We used 18 coarse rhetorical relations derived from 78 fine-grained ones. 
We used 40 documents in the training dataset as the development dataset, following 
\citet{kobayashi-etal-2022-simple}.

Instr-DT contains 176 documents obtained from the home-repair instruction manuals.
The number of rhetorical relations in the dataset is 39.
We followed Kobayashi et al.'s setting \citeyearpar{kobayashi-etal-2022-simple},
i.e., 
126, 25, and 25 documents were used for the training, development, and test datasets, respectively.

\textcolor{black}{
The GUM corpus contains 213 documents in total for 12 genres, e.g., News, Speech, Reddit, and Vlog. We used officially divided 165, 24, and 24 documents for training, development, and test datasets.
In this experiment, we translated rhetorical relation labels in the GUM corpus to match them with those in RST-DT by using a label correspondence described in \cite{liu-zeldes-2023-cant}.}

We used gold EDU segmentation for both datasets by following conventional studies.

\subsection{Evaluation Metrics \label{seq:metrics}}
We evaluated the results with micro-averaged F$_1$ scores of unlabeled, nuclearity-, relation-, and fully-labeled span, based on Standard-Parseval \cite{morey-etal-2017-much}, the standard evaluation metrics for RST discourse parsing.
Note that during both the training and test phases, RST-trees were converted into right-heavy binary trees \cite{sagae-lavie-2005-classifier}.

\begin{figure*}[t]
    \centering
    \includegraphics[width=\linewidth]{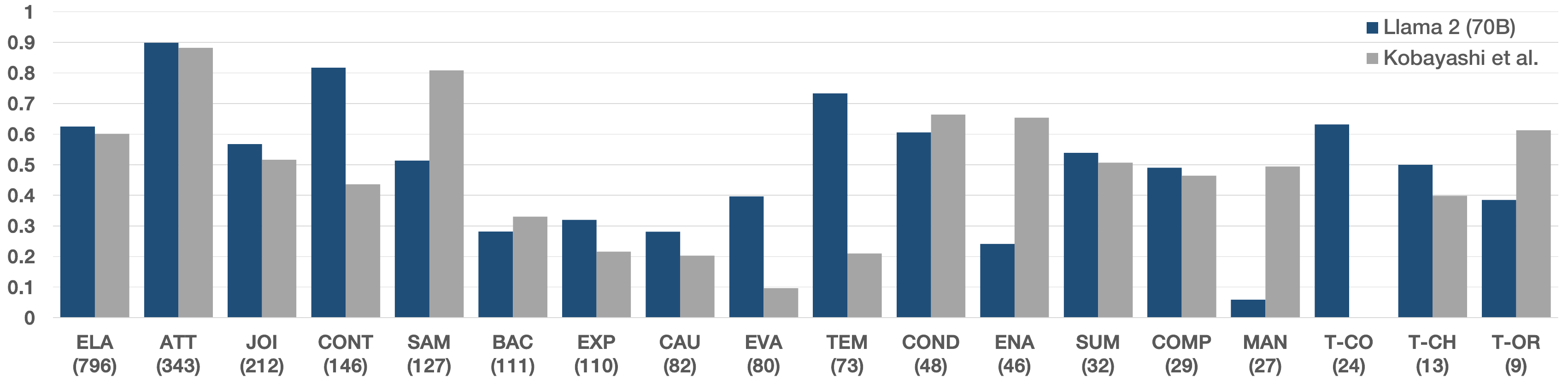}
    \caption{F$_1$ scores of Llama~2 (70B) and Kobayashi et al. for each relation label in bottom-up parsing: \textsc{ELAboration}, \textsc{ATtribution}, \textsc{JOInt}, 
    \textsc{CONTrast}, \textsc{SAMe-unit}, \textsc{EXPlanation}, \textsc{CAUse}, 
    \textsc{EVAluation}, \textsc{TEMporal}, \textsc{CONdition}, \textsc{ENAblement},
    \textsc{SUMmary}, \textsc{COMParison}, \textsc{MANner-means}, \textsc{Topic-Comment},
    \textsc{Topic-Change}, and \textsc{Textual-Organization}. Numbers in parenthesis are the frequency of the label.}
    \label{fig:rel}
\end{figure*}

\subsection{Configurations}

Our implementations are based on the official implementation of QLoRA,\footnotemark[2] which is based on Huggingface Transformers \cite{wolf-etal-2020-transformers}.
We employed Adam~\cite{kingma2015adam} to optimize the parameters.
We used QLoRA with 4-bit quantization, setting \texttt{lora\_r} to 64, \texttt{lora\_alpha} to 16, and \texttt{lora\_dropout} to 0.1.
A learning rate of 2e-4 was used
at a batch size of 16.
We scheduled the learning rate by linear warm-up, which increases it 
linearly during the first 3\% of training steps and then decreases it with cosine annealing to 0 until the final epoch.
We trained the model \textcolor{black}{with a different LoRA adapter for each subtask, i.e., span, nuclearity, and relation labeling, for} 5 epochs and 
chose the best checkpoint by evaluating the performance on the development dataset.
\textcolor{black}{We provide a summary of all hyperparameter settings in Appendix \ref{sec:hyperparameters}.}

\section{Results and Discussion}

\textbf{Overall Performance: }
\textcolor{black}{
Table \ref{tab:results} shows the results, where the scores of Kobayashi et al.'s and Yu et al.'s parsers are borrowed from their papers \cite{kobayashi-etal-2022-simple, yu-etal-2022-rst} and those of Guz et al's and Liu et al's parsers are borrowed from \cite{liu-zeldes-2023-cant}.
}
We show the results of our parsers with Llama~2 for 7B, 13B, and 70B parameters.

When focusing on the number of the parameters in Llama~2, 
the largest number naturally yields the best results.
In particular, 70B parameters obtained the current best scores for both datasets; however, the performance of the parsers with 7B and 13B parameters are still comparable to Kobayashi et al.'s parsers.
\textcolor{black}{
The gains by bottom-up parsing with 70B are impressive, surpassing Kobayashi et al.'s parser by approximately 3 points in RST-DT and Instr-DT, and by around 7 points in the GUM corpus.}

Bottom-up parsing consistently outperforms top-down parsing by 1 to 2 points when comparing the parsing strategies in our parsers. Since both strategies used the same prompts for nuclearity and rhetorical relation labeling, the differences are considered to come from different prompts to build the skeleton of the RST tree. The prompts for bottom-up parsing mention three text spans in \texttt{Stack1}, \texttt{Stack2}, and \texttt{Queue1}, while those for top-down parsing mention only one text span in \texttt{Input}.
In other words, the former handles richer information than the latter. 

\noindent\textbf{Performance of Relation Labeling}:
To investigate the effectiveness of our parsers in more detail, we show Relation F$_1$ scores for each relation label in RST-DT with Llama~2 (70B) and Kobayashi et al.'s bottom-up parsers in Figure \ref{fig:rel}. 
In most cases, Llama~2 (70B) outperforms Kobayashi et al., even for less frequent relation labels, indicating Llama~2 (70B) has greater potential for generalization.

These results indicate the effectiveness of LLMs for RST discourse parsing. The findings are interesting in that simple pre-trained language models consisting of only a transformer decoder can be easily tailored for determining parsing actions by fine-tuning with prompts. 
Our parsers perform well on all datasets, 
which are from different domains.
The results demonstrate the advantage of LLM-based RST discourse parsing in domain portability, particularly in achieving the best scores on Instr-DT with less training data.

\noindent\textbf{Cross-corpus Generalization}:
To examine the generalizability of our parsers in detail, we evaluated them with \textcolor{black}{out-of-domain evaluations using RST-DT and the GUM corpus.}
Tables \ref{tab:RSTDT2GUM} and \ref{tab:GUM2RSTDT} show the results of training parsers on one dataset and evaluating them on the other.

\textcolor{black}{
Comparing the results with Table \ref{tab:results}, the performances are lower. 
In particular, the parsers trained with RST-DT degraded more when tested on the GUM corpus than the opposite, 
the parsers trained with the GUM corpus and tested on RST-DT.
The findings suggest that using a single genre dataset for generalization across multiple genres is challenging. This aligns with the observation made by \citet{liu-zeldes-2023-cant}.
}

\textcolor{black}{
In Table \ref{tab:RSTDT2GUM}, when comparing our parsers with Liu et al.'s and Kobayashi et al.'s parsers, our parsers obtained better scores than them in most cases. 
In particular, Llama~2 (70B) with the bottom-up strategy achieved the best scores. It obtained around 2-point gains against Kobayashi et al.'s parser in all metrics.
Notably, it further outperformed both Liu et al.'s and Guz et al.'s parsers trained with the GUM corpus in Table \ref{tab:results} on Span while it obtained a slightly lower score in Nuc.
}


\textcolor{black}{
On the other hand, performance degradation in Table \ref{tab:GUM2RSTDT} is much lower than that in Table \ref{tab:RSTDT2GUM}. Our parsers obtained remarkable gains against Liu et al.'s and Kobayashi et al.'s parsers. The gains are emphasized in Rel. and Full.
Despite being trained with an out-of-domain dataset, our parsers in Table \ref{tab:RSTDT2GUM} achieved Span F$_1$ scores that are comparable to those in Table \ref{tab:results}.
Furthermore, the performance of our parsers in Full degraded by 12\%, 
while that of Kobayashi et al.'s parser degraded by 15-20\%.
}

\textcolor{black}{
The successful improvements were achieved by using a large number of LLM parameters and training them with a massive amount of text, compared to the encoder-only models.
Although we could potentially improve the performance of the encoder-only PLMs by increasing their parameters to the level of the current LLMs, this task poses a significant challenge. Furthermore, considering the focus of the research has shifted from encoder-only PLMs to LLMs, our findings are highly valuable for future research in RST discourse parsing.
}

\renewcommand{\arraystretch}{0.9}
\begin{table}[t]
    \centering
    \small
    \begin{tabular}{llcccc}
    \toprule
    & & \textbf{Span} & \textbf{Nuc.} & \textbf{Rel.} & \textbf{Full} \\
    \midrule
    \multirow{5}{*}{\rotatebox[origin=c]{90}{Top-down}}
    & Liu et al.  & 66.2 & 50.8 & $-$ & $-$\\
    & Kobayashi et al. & 70.3& 53.7& 41.1&38.3\\
    & Llama~2 (7B) & 68.2 & 51.8 & 39.8 & 37.6 \\
    & Llama~2 (13B) & 69.0 & 51.5 & 39.2 & 37.3 \\
    & Llama~2 (70B) & 71.0 & 53.3 & 42.1 & 39.8 \\
    \midrule
     \multirow{5}{*}{\rotatebox[origin=c]{90}{Bottom-up}}
    & Guz et al. & 65.3 & 49.5 & $-$ & $-$\\
    & Kobayashi et al. & 68.3& 52.6& 40.8&38.0\\
    & Llama~2 (7B) & 69.4 & 53.0 & 40.2 & 38.2 \\
    & Llama~2 (13B) & 69.3 & 52.1 & 39.8 & 37.8 \\
    & Llama~2 (70B) & 72.6 & 55.6 & 43.0 & 40.5 \\
    \bottomrule
    \end{tabular}
    \caption{Cross-corpus generalization results on the GUM corpus. 
    \textcolor{black}{
    Parsers were trained using RST-DT and their performance was evaluated on the GUM corpus.}
    }
    \label{tab:RSTDT2GUM}
\end{table}

\renewcommand{\arraystretch}{0.9}
\begin{table}[t]
    \centering
    \small
    \begin{tabular}{llcccc}
    \toprule
    & & \textbf{Span} & \textbf{Nuc.} & \textbf{Rel.} & \textbf{Full} \\
    \midrule
    \multirow{5}{*}{\rotatebox[origin=c]{90}{Top-down}}
    & Liu et al.  & 72.7 & 57.4 & $-$ & $-$\\
    & Kobayashi et al.  & 76.1& 61.8& 49.3&46.5\\
 & Llama~2 (7B) & 75.3 & 61.7 & 49.9 & 48.0 \\
    & Llama~2 (13B) & 76.3 & 63.4 & 51.3 & 49.4 \\
    & Llama~2 (70B) & 78.2 & 64.4 & 51.5 & 49.7 \\
    \midrule
     \multirow{5}{*}{\rotatebox[origin=c]{90}{Bottom-up}}
    & Guz et al. & 71.1 & 55.9 & $-$ & $-$\\
    & Kobayashi et al. & 72.0& 58.5& 46.3& 44.3\\ 
    & Llama~2 (7B) & 77.4 & 63.6 & 51.3 & 49.0 \\
    & Llama~2 (13B) & 77.4 & 64.5 & 52.2 & 50.3 \\
    & Llama~2 (70B) & 79.7 & 66.5 & 53.2 & 51.1 \\
    \bottomrule
    \end{tabular}
    \caption{Cross-corpus generalization results on RST-DT. 
    \textcolor{black}{
    Parsers were trained using the GUM corpus and their performance was evaluated on RST-DT.}
    }
    \label{tab:GUM2RSTDT}
\end{table}

\section{Conclusion}
This paper explored the potential of using Llama~2, the largest publicly available decoder-only language model, pre-trained with a trillion tokens of text data, for RST discourse parsing. 
To exploit Llama~2, we translated fundamental bottom-up and top-down parsing processes into prompts. 
Then, we fine-tuned Llama~2 with them using QLoRA for efficient computing.
The experimental results obtained from three datasets, RST-DT, Instr-DT, and the GUM corpus, which come from different domains, demonstrated the effectiveness of our parsers with Llama~2, including their domain portability.
\textcolor{black}{
Specifically, our approach with Llama~2 (70B) obtained better results than the current SOTA parser on all datasets. Furthermore, findings from the experimental results for cross-corpus generalization showed the significant promise of our approach, that is, that our parsers, in spite of being trained with the GUM corpus, obtained comparable performance to parsers trained with RST-DT itself in Span F$_1$ scores, keeping small degradation in Nuc., Rel., and Full F$_1$ scores. 
}

\textcolor{black}{
Since this work is just a first step in exploiting LLMs for RST discourse parsing, we can work in this direction with various topics to further improve the LLM-based model; e.g., incorporating richer information to improve the current top-down parsing model.
}

\subsubsection*{Limitations}
While our parsers achieved the best performance with high generalizability, they have serious limitations: Our parsers require a significant amount of computational resources and time.
Although our parsers with Llama~2 (7B) work on a standard GPU with 24~GB memory, such as RTX 3090, and those with 13B require over 40~GB GPU memory, those with 70B require a high-end GPU with 80~GB memory, such as A100. 
To train the parsers, we need 1-2 days for 7B and 13B; however, we need nearly five days for 70B.
Furthermore, it takes several minutes to parse a document with our models, whereas it takes only a few seconds with a standard parser. 

\section*{Acknowledgements}
Part of this work was supported by JSPS KAKENHI Grant Numbers P21H03505.     
\bibliography{anthology,custom}

\appendix

\clearpage

\section{Hyperparameters}
\label{sec:hyperparameters}

\begin{table}[t]
    \centering
    \small
    \begin{adjustbox}{width=\linewidth}
    \begin{tabular}{cc}
        \toprule
        \multicolumn{2}{c}{\textbf{Computing Interface}} \\
        \midrule
        Experiments for Llama~2 (7B) & NVIDIA RTX 3090 \\
        \midrule
        Experiments for Llama~2 (13B) & NVIDIA RTX A6000 \\
        \midrule
        Experiments for Llama~2 (70B) & NVIDIA A100 (80~GB) \\
        \midrule
        \multicolumn{2}{c}{\textbf{Hyperparameters}} \\
        \midrule
        number of training epochs & 5 \\
        \midrule
        batch size & 16 \\
        \midrule
        optimizer & Adam \\
        \midrule
        learning rate & 2e-4 \\
        \midrule
        learning rate scheduler & \begin{tabular}{c}Linear warm-up and \\cosine annealing\end{tabular} \\
        \midrule
        warm-up ratio & 0.03 \\
        \midrule
        gradient clipping & 1.0 \\
        \midrule
        lora $r$ & 64 \\
        \midrule
        lora $\alpha$ & 16 \\
        \midrule
        lora dropout ratio & 0.1 \\
        \midrule
        lora target modules & \begin{tabular}{c}All linear layers in \\transformer-blocks\end{tabular}\\
        \midrule
        quantization for Llama~2 & \begin{tabular}{c}4-bit NormalFloat and \\double quantization\end{tabular} \\
        \bottomrule
    \end{tabular}
    \end{adjustbox}
    \caption{Hyperparameters in the experiments}
    \label{tab:hyperparameters}
\end{table}

Table \ref{tab:hyperparameters} shows the hyperparameters and computing interfaces used in our
experiments.

\end{document}